\documentclass[final]{cvpr}
\usepackage{booktabs,subcaption,amsfonts,dcolumn}
\usepackage{comment}
\usepackage{times}
\usepackage{epsfig}
\usepackage{amsmath}
\usepackage{amssymb}
\usepackage{algorithm}
\usepackage{algpseudocode}
\usepackage{booktabs}
\usepackage{subcaption}
\usepackage{graphicx}
\usepackage{enumitem}
\usepackage{scrextend}
\usepackage{soul}
\usepackage[title]{appendix}
\usepackage[pagebackref=true,breaklinks=true,colorlinks,bookmarks=false]{hyperref}
\usepackage{xcolor}

\def\cvprPaperID{7001} 
\def\confYear{CVPR 2022}

\begin{document}

\title{ISNAS-DIP: Image-Specific Neural Architecture Search for Deep Image Prior}


\author{Metin Ersin Arican$^1$\thanks{equal contribution}\,, Ozgur Kara$^1$\footnotemark[1]\,,
Gustav Bredell$^2$\,,
Ender Konukoglu$^2$\\
{\small$^1$Department of Electrical and Electronics Engineering, Bogazici University, Istanbul, Turkey}\\
{\small$^2$ Department of Information Technology and Electrical Engineering, ETH-Zurich, Zurich, Switzerland}\\
{\tt\small$^1$\{metin.arican, ozgur.kara\}@boun.edu.tr}\\
{\tt\small$^2$\{gustav.bredell, ender.konukoglu\}@vision.ee.ethz.ch}}

\maketitle

\begin{abstract}
   Recent works show that convolutional neural network (CNN) architectures have a spectral bias towards lower frequencies, which has been leveraged for various image restoration tasks in the Deep Image Prior (DIP) framework. 
   The benefit of the inductive bias the network imposes in the DIP framework depends on the architecture.  
   Therefore, researchers have studied how to automate the search to determine the best-performing model. 
   However, common neural architecture search (NAS) techniques are resource and time-intensive. 
  Moreover, best-performing models are determined for a whole dataset of images instead of for each image independently, which would be prohibitively expensive. 
  In this work, we first show that optimal neural architectures in the DIP framework are image-dependent.
  Leveraging this insight, we then propose an image-specific NAS strategy for the DIP framework that requires substantially less training than typical NAS approaches, effectively enabling image-specific NAS.
   We justify the proposed strategy's effectiveness by (1) demonstrating its performance on a \textit{NAS Dataset for DIP} that includes 522 models from a particular search space (2) conducting extensive experiments on image denoising, inpainting, and super-resolution tasks. Our experiments show that image-specific metrics can reduce the search space to a small cohort of models, of which the best model outperforms current NAS approaches for image restoration. Codes and datasets are available at \href{https://github.com/ozgurkara99/ISNAS-DIP}{https://github.com/ozgurkara99/ISNAS-DIP}.

\end{abstract}

\section{Introduction}

Convolutional neural networks (CNNs) have been ubiquitously utilized in almost every field of computer vision. Particularly, researchers harness the power of CNNs in image restoration tasks \cite{Zhang2017BeyondAG, Zhang2017LearningDC, Zhang2018FFDNetTA, Guo2019Cbdnet}, which refers to the task of recovering the original image from a corrupted version. The success of CNNs comes as a result of their ability to learn a mapping from a corrupted image to its uncorrupted counter-part.
However, the ground truth labels are not always available to learn such a mapping for a given domain, limiting the use of approaches under supervised settings. To tackle this problem, researchers orient their attention towards unsupervised approaches. Recent discoveries have shown that the architecture of CNNs contains an intrinsic \textit{prior} that can be used in image restoration tasks \cite{Ulyanov_2018_CVPR, Saxe2011OnRandom}. This insight led to the \textit{Deep Image Prior} (DIP) framework \cite{Ulyanov_2018_CVPR}, which works solely with the degraded image and can produce competitive results for image restoration tasks without a supervised training phase. It offers an alternative solution to restoration problems by suggesting a new regularizer: the network architecture itself. In addition to this empirical discovery, Rahaman \etal~\cite{pmlr-v97-rahaman19a} investigated the spectral bias of neural networks towards low frequencies theoretically, which can explain the impressive performance of the DIP framework. Chakrabarty~\cite{Chakrabarty2019TheSB} further explored the underlying reason behind the success of DIP in denoising natural images. The work demonstrates that the network tends to behave similarly to a low pass filter at the early stages of iterations. Finally, DeepRED~\cite{Mataev2019DeepREDDI} merged the concept of ``Regularization by Denoising" (RED) by adding explicit priors to enhance DIP. 

One problem faced in the DIP framework is that the architectural design of the network has a substantial impact on the performance. Recent works attempted to automate the search process of network architecture for various tasks, which is referred to as the \textit{Neural Architecture Search} (NAS). In the context of DIP, Chen \etal~\cite{Chen2020NAS-DIP} applied NAS to the DIP framework. However, current NAS approaches come with substantial computational costs, as they require optimizing a large number of architectures to determine the optimum. Moreover, this cost prohibits determining the optimum architecture for every image; instead, existing NAS approaches search for the best architecture for a dataset of images.

\newpage
\noindent
\textbf{Our work.} 
In this paper, we propose novel image-dependent metrics to determine optimal network architectures in the DIP framework with minimal training. Unlike previous works, we apply our metrics to DIP for finding image-specific architectures since performance is strongly dependent on the content of the image that is to be restored.

\begin{figure}[ht]
     \begin{subfigure}[b]{0.5\linewidth}
         \includegraphics[width=\linewidth]{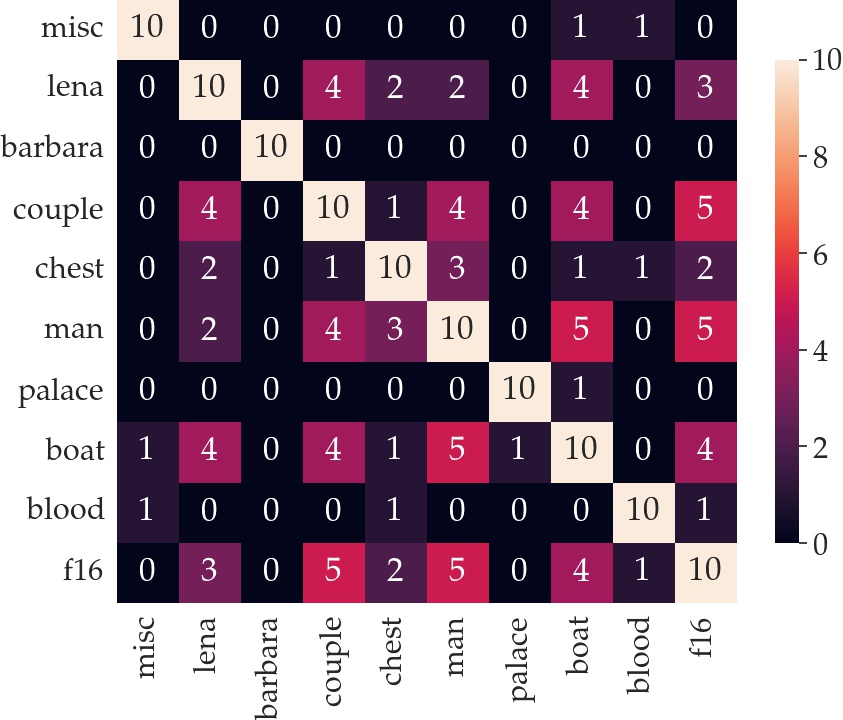}
         \caption{Denoising}
         \label{fig:dn overlap}
     \end{subfigure}%
     \begin{subfigure}[b]{0.5\linewidth}
         \includegraphics[width=\linewidth]{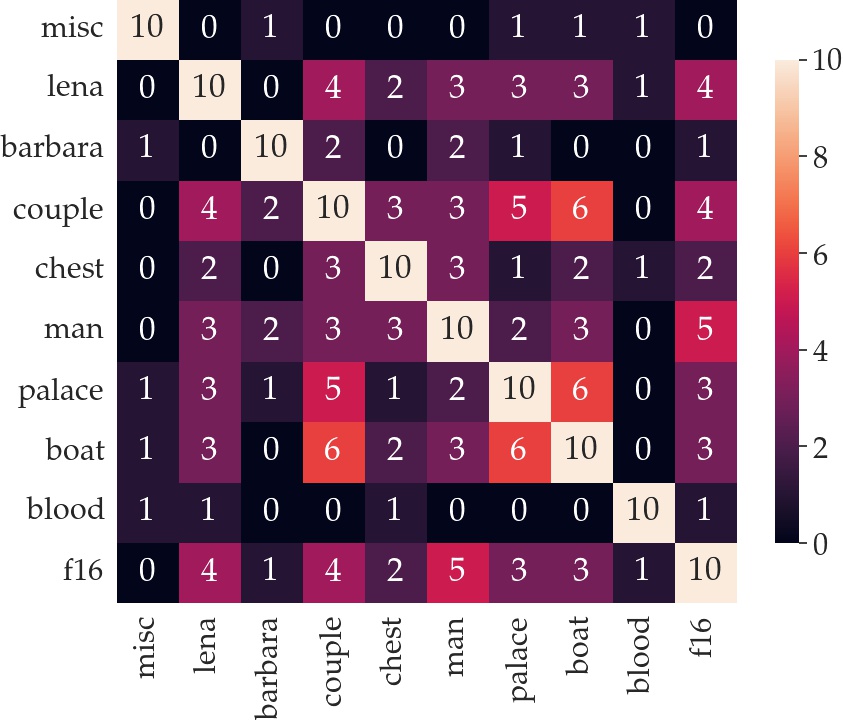}
         \caption{Inpainting}
         \label{fig:in overlap}
     \end{subfigure}
        \caption{The overlap of the best-performing architectures between different images is shown here. The numbers indicate how many of the top-10 models of the search space for image x are also in the top-10 for image y. This is shown for the task of denoising and inpainting in (a) and (b), respectively. E.g. the value at the intersection of chest and lena in the denoising heatmap, which is 2, indicates that there are 2 models in each of the images' best-performing 10 models that are the same.}
        \label{fig:overlap}
\end{figure}

We first motivate image-specific NAS, by showing that in a given search space, 
there is only a small overlap of the best architectures for different images. This is illustrated in Figure~\ref{fig:overlap}, where the matrices show the number of overlaps between the top 10 models (of a total of 522 models) for each image for denoising and inpainting.

To identify architectures that are fitting for a specific image, we propose image-dependent metrics that measure the property of how far the power spectral density (PSD) of the generated initial output of a network is from that of the corrupted image and use it as our metric. The intuition relies on the fact that the more these two are similar, the better the model will reconstruct the image since it is closer to the solution space. 

We motivate the choice of metrics by looking at the correlation between the metrics' values and image restoration performance. 
There is an imperfect correlation; hence we select a small cohort of architectures to optimize based on the metrics' values. 
A final selection is then made by selecting the model whose output is closest to the average of outputs of all models.

We conduct experiments on conventional datasets for image denoising, image inpainting, and single image super-resolution tasks using the proposed strategy. For each image in the datasets, we run our ISNAS algorithm to identify the optimal image-specific models. The results demonstrate that our method is superior to the state-of-the-art work~\cite{Chen2020NAS-DIP} in terms of its quality improvement. 

\textbf{The main contributions} can be summarized as follows:
\begin{itemize}[noitemsep]
    \item We empirically show the necessity of identifying image-specific models to augment the quality of DIP.
    \item We present novel metrics to be used in NAS requiring only the randomly initialized CNN network. These metrics allow ranking architectures within any search space without lengthy optimization as a surrogate to their success on image restoration tasks.
    \item We introduce two selection procedures among a subset of models for finding optimal architectures in an unsupervised fashion for DIP.
    \item We generate a \textit{NAS Dataset for DIP} having 522 models optimized for ten images from different domains, including image denoising and image inpainting tasks,.
    \item Extensive experiments on commonly used datasets and \textit{NAS Dataset for DIP} validate our approach.

\end{itemize}

\section{Related Work}

\subsection{NAS}
The challenges of designing complex architectures by hand have shifted researchers' interest to the area of automatic neural architecture search \cite{Elsken2019NeuralAS}. One of the initial works \cite{Zoph2017NeuralAS} formulated NAS as a reinforcement learning problem, where better architectures are sampled by training a policy network. Building upon this work, Zoph \etal~\cite{Zoph2018LearningTA} applied a cell-based search which is then stacked to constitute the network. Pham \etal~\cite{pmlr-v80-pham18a} introduced the idea of \textit{weight sharing} where the networks are trained jointly. They show that it reduces the complexity by 1000 times. In the context of image restoration, Suganuma \etal~\cite{suganumaICML2018} employed evolutionary search to convolutional autoencoders. Subsequently, Zhang \etal developed HiNAS~\cite{Zhang2020MemoryEfficientHN}, which utilized gradient-based search strategies and introduced a hierarchical neural architecture search for image denoising. 
A recent study \cite{Chen2020NAS-DIP} utilized a reinforcement learning-based NAS to improve the network architecture and offered a search space for upsampling blocks, of which we follow the same procedures throughout our experiments. 

\begin{figure*}[ht]

\centering
\includegraphics[width=0.90\textwidth]{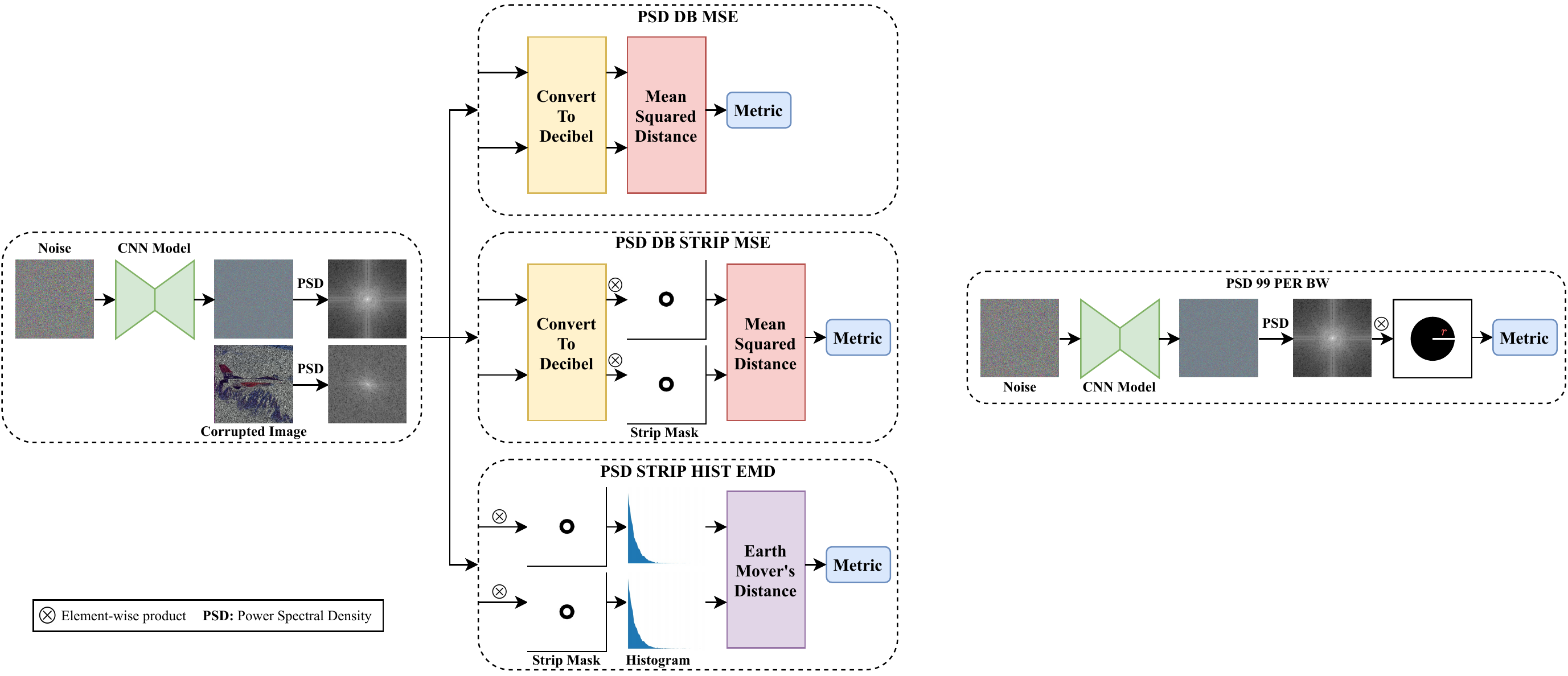}
\caption{Visualization of proposed metrics. The left block with three branches demonstrates the image-dependent metrics. All of them use the PSDs of corrupted images and random output. The right block shows the image-independent metric, which only requires a randomly initialized CNN architecture}
\label{fig:metric summary}
\end{figure*}

\subsection{NAS Without Training}
One of the major challenges in the NAS is its computational expense since numerous networks need to be evaluated before determining the best model. To minimize resource consumption, researchers have proposed \textit{training-free} approaches. Recently, Mellor \etal~\cite{Mellor2021NeuralAS} proposed to assign scores to network architectures at initialization requiring no training. Their intuition is that the closeness of binary codes associated with two inputs is an indicator of how much the network is able to separate these. They empirically show the correlation between the initial score and the network's performance, although it lacks theoretical explanations. Chen \etal~\cite{chen2020tenas} attempted to bridge the deep learning theory and NAS applications by questioning and analyzing the effectiveness of two indicators on ranking the networks prior to training, namely \textit{trainability}, and \textit{expressivity}. Zhang \etal~\cite{Zhang2021DifferentiableAS} went a step further and proposed a metric based on \textit{synaptic saliency} that does scoring requiring neither training nor labels. They achieved promising results on NAS-Bench-201 \cite{Dong2020NASBench201ET}, which is a conventional benchmark used for evaluating NAS algorithms. However, existing NAS benchmarks are mainly composed of image classification datasets.

Inspired by the previous \textit{training-free} approaches, we adapt the idea of ranking networks at initialization to DIP settings. Furthermore, to evaluate our NAS approach, we generate our \textit{NAS Dataset} for DIP analogous to NAS-Bench 201 \cite{Dong2020NASBench201ET}, which is explained in Section \ref{4_1}.

\section{Methodology}

In a typical NAS algorithm, the main bottleneck, in terms of time, is the training of the models to compute their performance. If one can find an easy-to-calculate training-free performance predictor, this bottleneck can be eliminated. In addition, our experiments show that model selection for DIP settings should be image-dependent. In this paper, we propose several different training-free and image-dependent performance predictors and study their effectiveness.


Ulyanov \etal~\cite{Ulyanov_2018_CVPR} observed that the architectures with better performances in DIP tend to have outputs possessing large spatial structures at the early iterations of training. One useful metric to capture the distribution of the spatial structures is to use power spectral density (PSD). Coarse and fine textures will lead to a PSD that is concentrated on low and high frequencies, respectively.  


Inspired by these, we put forward a hypothesis that if the PSD of an untrained CNN's output is \textit{similar} to that of the image to be reconstructed, then the model will be closer to the desired solution space, hence it will facilitate the optimization and lead to better restoration results compared to others with lower similarity. In this section, we formulate different metrics to quantify the similarity between an image and the CNN's random output.



\subsection{Image Dependent Metrics}



It would be preferable to compute the distance using the ground truth image. However, in a practical situation, we do not have access to them. Therefore, we use the distance between the CNN's random output and the corrupted image as a proxy for the distance between the output and the ground truth image.

\subsubsection{PSD DB MSE}

One straightforward approach that can be used to measure the distance between PSDs is the mean square error (MSE). Generally, the PSD of an array consists of numbers that are orders of magnitudes different from each other. Thus, in MSE calculation, using the logarithm of the PSD, which we call decibel PSD,  instead of directly PSD itself is better suited. To that end, we first calculate the decibel PSD's of the output of a given CNN with randomly initialized weights and the corrupted image. Then, we measure the MSE between them. A schematic representation of this metric can be seen in Fig. \ref{fig:metric summary} and it is formulated as
\begin{equation}
    \frac{1}{n} \sum_{i,j} \left( 10\cdot \log X_{i,j} - 10 \cdot \log Y_{i,j} \right)^2
\end{equation}
where $X$ and $Y$ denote the corresponding power spectral densities of the CNN's random output and a corrupted image, respectively, and $n$ denotes the number of pixels.

\subsubsection{PSD DB Strip MSE}

Spatial structures and texture in an image are related to its PSD, but each frequency region of the PSD does not equally contribute to the spatial structures. 
Very high-frequency regions of the PSD of a corrupted image are heavily affected by noise. Hence, focusing on only a band of frequencies around the center in similarity comparison is more suitable.
In light of these insights, the metric is calculated as follows: First, a mask is applied to the decibel of the PSDs of the CNN's random output and the corrupted image. Then, we calculate the MSE between them. As the mask, we employ a strip having inner and outer diameter sizes of 10\% and 20\% of the image size, respectively, to reduce the dependency on the image size. A schematic representation of this metric can be seen in Fig. \ref{fig:metric summary} and it is formulated as
\begin{equation}
    \frac{1}{n} \sum_{i,j} \left( 10 \cdot \log X_{i,j} - 10 \cdot \log Y_{i,j}  \right)^2 \cdot M_{i, j}
\end{equation}
where $X$ and $Y$ denote the corresponding power spectral densities of the CNN's random output and a corrupted image, respectively, $M$ denotes the mask, a 2-D array of 1s and 0s, and $n$ denotes the number of non-zero pixels of the mask $M$.

\subsubsection{PSD Strip Hist EMD}

To make our metrics rotation invariant, we use the histograms of PSDs. 
In this metric, we first calculate the PSDs of the CNN's random output and a corrupted image. Then, we discard the entries of PSDs, where the corresponding entry of a mask is zero. For this, we use the mask defined in the previous metric (PSD DB Strip MSE). Afterwards, the PSDs are flattened into two 1-D arrays, which are then converted to histogram representations. Finally, we calculate the earth mover's distance (EMD) between these two histograms. The range and number of bins of the histograms are determined as a result of trials. In our experiments, we use 75 as the number of bins and 0-1 as the range of the histograms.  A schematic representation of this metric can be seen in Fig. \ref{fig:metric summary}.

\subsection{Image Independent Metrics}

In our evaluation, we also included an image independent metric also using the PSD, inspired by the structural bias of CNNs as exploited by Heckel~\cite{Heckel2020DenoisingAR}. 
This allows us to dissect whether the contribution is due to using the PSD of the CNN-generated image or the image dependency of the metrics described above.
The structural part of an image can be thought of as the low-frequency component, just as noise or corruptions are of high-frequency components. Relying on the hypothesis that if the frequency spectrum of the output of a randomly initialized CNN is concentrated on low-frequency regions, then it tends to perform better in restoration tasks such as denoising and inpainting, we propose a metric to measure the low-pass characteristic of a CNN and use it as our image independent metric.

\subsubsection{99\% Bandwidth (99 BW)}

A straightforward method to quantify the low-frequency nature of an array is to calculate its bandwidth. We define the $P\%$ bandwidth of a 2D array as the radius of the circle containing the $P\%$ of the total energy in the PSD of the 2D array.

Obviously, the bandwidth of a CNN does not depend on the image to be reconstructed, which makes it an image independent metric. To choose a value of P, we created several outputs from randomly initialized CNNs and selected the P value resulting in the most variation in bandwidths. This was $P = 99$, so we used this value in our experiments.






\subsection{ISNAS-DIP Overview}
Given any search space, 
we can use one of the metrics to calculate the fitness score of models to shrink the entire search space into a few models. 
As mentioned, the metrics are correlated to the performance of the restoration of an optimized network, but the correlation is imperfect.
Therefore, we first calculate the metrics of the models at initialization, i.e., without any optimization of the weights. Then, we sort each model in ascending order for ``Image Independent" and ``Image Dependent" metrics according to their values. Finally, for each metric, we choose the top-N models having the lowest metric values. 

Next, we have to make a final selection between the chosen N models. We propose two \textit{averaging techniques} to perform the selection. The first selection procedure is as follows: Optimize the chosen N models under DIP settings. Take the average of the N reconstructed images that have coefficients inversely proportional to the value of the corresponding models' metrics. Calculate the MSE score between the average and each reconstructed image. Select the model that corresponds to the reconstructed image giving the lowest MSE with average restored, that is to say, closest one to the average. We will call this technique as \textit{full-sized averaging selection} throughout the paper.

The second technique differs in the size of the image to be optimized by each model and dramatically increase the speed of ISNAS. Before running optimization for N models, we resize the corrupted image to ``$64\times64$" by rescaling to speed up the optimization process. We will call this technique \textit{resized averaging selection} throughout the paper. 
By choosing the model whose output is closest to the average, we at least guarantee to exclude the worst models among N models. Note that these selection methods introduce a negligible image dependency for 99 BW metric since the final model selection among N models may depend on the content of the image.  






\section{Experimentation}
In this section, we first describe the experiment set-up, including the search space, the datasets that we use, and the implementation details. We then continue with the analysis of the architecture selection on a small dataset, composed of both natural and medical images, denoted as \textit{NAS Dataset for DIP}, and further evaluate image restoration performance for denoising, inpainting, and super-resolution on established datasets for the given restoration task.


\subsection{Architecture Selection in NAS Dataset}\label{4_1}
\subsubsection{Experiment Set-up}
\noindent
\textbf{Search Space.}
Following the recent works, we use the same search space as NAS-DIP~\cite{Chen2020NAS-DIP} throughout our experiments. The search space consists of different types of upsampling cells and random cross-level feature connections between decoder and encoder cells. Each upsampling cell is defined by five discrete attributes: Spatial feature sampling, feature transformation, kernel size, dilation rate, activation layer. If the connection scale factors are higher than $2\times$, the series of $2\times$ upsampling operations are connected consecutively (e.g for $4\times$ connection between the decoder and encoder cell, two consecutive 2x upsampling operations are connected). Following the original work \cite{Chen2020NAS-DIP}, each architecture is constituted of 5 encoder and 5 decoder cells. In total, we randomly selected 522 different models in the search space and recorded each model's results for image denoising and image inpainting tasks.

\begin{figure}[ht]
\centering
\includegraphics[width=\linewidth]{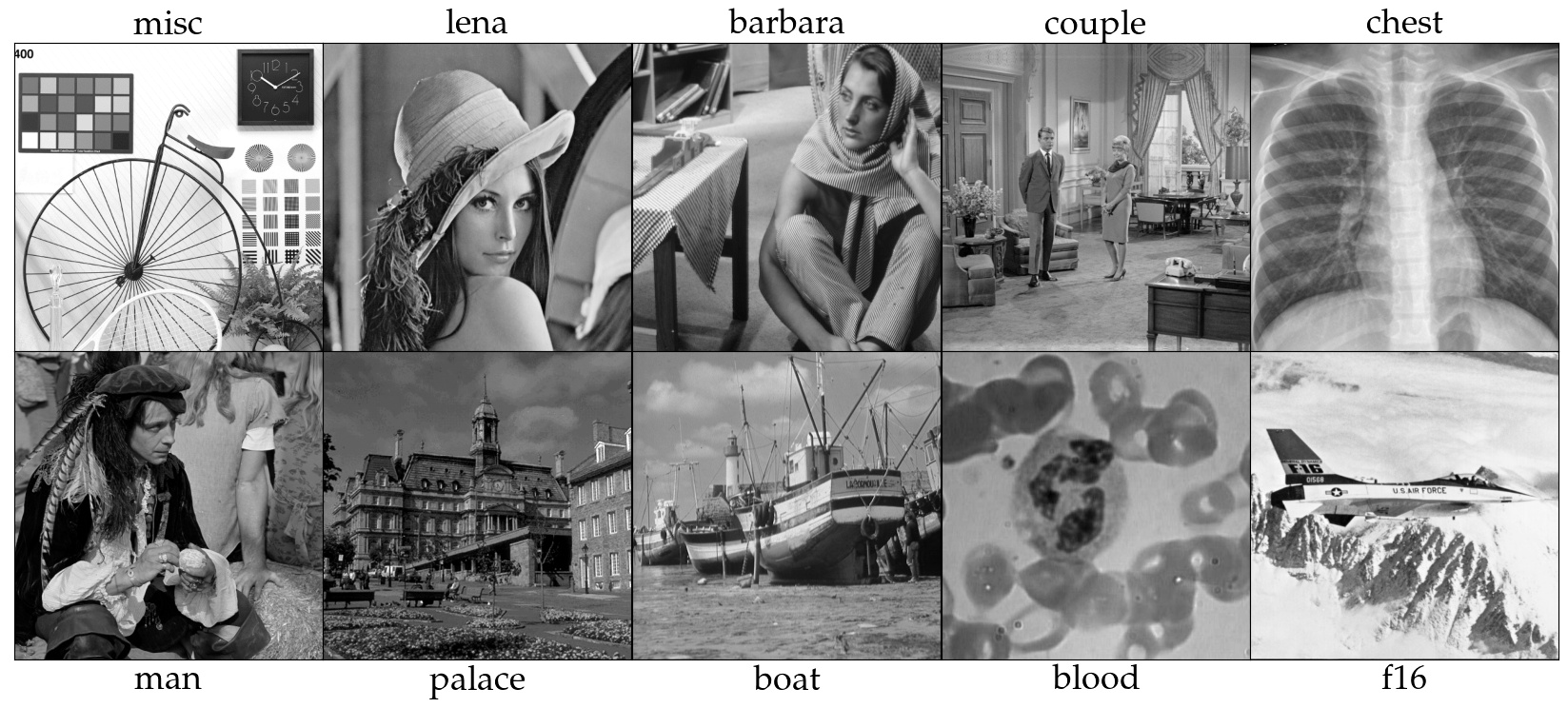}
\caption{Images in \textit{NAS Dataset for DIP}. Images are selected from BM3D Dataset~\cite{Dabov2007VideoDB}, Dataset of Standard $512\times512$  Grayscale  Test  Images\textsuperscript{1} , BCCD dataset\textsuperscript{2}, and Chest X-Ray Images (Pneumonia)~\cite{chestDataset}.}
\label{fig:nas_dataset}
\end{figure}

\noindent
\textbf{NAS Dataset for DIP.}
To evaluate how well our metrics perform a model selection in the search space, we created a \textit{NAS Dataset for DIP} using the images in Fig \ref{fig:nas_dataset}. 

The dataset is composed of 10 images, 8 are selected from either ``BM3D Dataset~\cite{Dabov2007VideoDB}" or ``Dataset of Standard $512\times512$ Grayscale Test Images"\footnote{\label{note1}\href{https://ccia.ugr.es/cvg/CG/base.htm}{https://ccia.ugr.es/cvg/CG/base.htm}}. The images ``blood"\footnote{\label{note2} Taken from \href{https://github.com/Shenggan/BCCD_Dataset}{Link} and is under MIT licence} and ``chest"~\cite{chestDataset} are appended to offer domain diversity. Each image is converted to grayscale and resized to $512\times512$. Afterwards, for the image denoising task, Gaussian noise with $\sigma=25$ is applied; for image inpainting, a Bernoulli mask with 50\% is applied. Analogous to NAS-Bench 201 \cite{Dong2020NASBench201ET}, 
\textit{NAS Dataset for DIP} allows us to measure the capability of ISNAS algorithm. We optimized the 522 models for all the generated images in this dataset for both denoising and inpainting tasks. 



\noindent
\textbf{Implementation Details.} While training the models, 
we use Adam as an optimizer with a constant learning rate of 0.01. We choose the stopping point as 1200 for denoising, 9500 for inpainting, and 4500 for super-resolution.

We use the final PSNR score as our evaluation metric and follow the same procedure with DIP~\cite{Ulyanov_2018_CVPR}, where an exponential sliding window is applied to the resulting images. 


\subsubsection{Architecture Selection Accuracy}

Using the \textit{NAS Dataset for DIP} and the 522 models, we empirically determined that the same model architectures behave differently to different images under the DIP setting (see Figures~\ref{fig:dn overlap} and~\ref{fig:in overlap}, and Appendix for further results). 

\begin{figure*}[ht]
\centering
     \includegraphics[width=\textwidth]{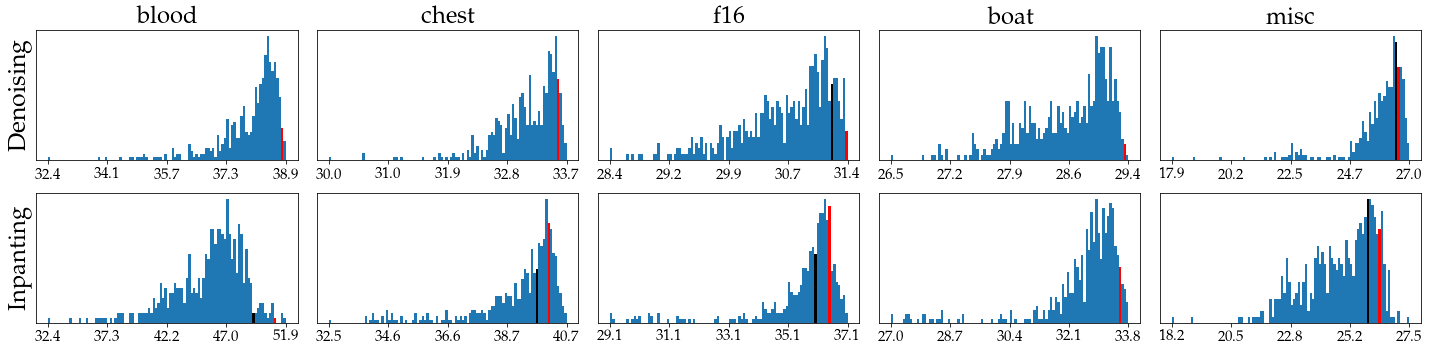}
    
\caption{Histograms of the final PSNR scores of models for five images in \textit{the NAS Dataset for DIP} for image denoising (first row) and image inpainting (second row). Non-blue colored bars represent the final PSNR scores of the best models included in top-N models found by PSD DB Strip MSE metric. The bin depicted with \textcolor{red}{red} denotes the best model among N=15 models, the black bin denotes the best one among N=5 models. Histograms of the remaining images and random selection results are provided in appendix.}
\label{fig:histogram}
\end{figure*}


Building on the insight that image restoration tasks benefit from image-specific neural network architectures, we investigate how accurate the suggested metrics are at identifying the best network architecture for the specific image. Fig. \ref{fig:histogram} shows the distribution of the final PSNR scores of the 522 models for both image denoising (first row) and image inpainting (second row). We observe that the found model by PSD DB Strip MSE with N=15 (the best-performing one in 15 models is depicted with \textcolor{red}{red} bin) is 
always located on the right side of the peak value for all images. In addition, the \textcolor{black}{black} bin shows the same score for N=5 case. Even though the latter case is worse than N=15, it is still competitive for some cases. Thus, PSD DB Strip MSE has the potential to reduce the size of the search space into only a few models. 

\begin{table}[h]
\centering
\resizebox{\linewidth}{!}{
\begin{tabular}{lccccccc}
\toprule
\multicolumn{1}{c}{Images} &  \multicolumn{2}{c}{\begin{tabular}[c]{@{}c@{}}PSD DB \\ MSE\end{tabular}} & \multicolumn{2}{c}{\begin{tabular}[c]{@{}c@{}}PSD DB \\ STRIP MSE\end{tabular}} & \multicolumn{2}{c}{\begin{tabular}[c]{@{}c@{}}PSD STRIP\\ HIST EMD\end{tabular}} & \multicolumn{1}{c}{\begin{tabular}[c]{@{}c@{}}PSD 99 PER\\ BW\end{tabular}} \\ 
\cmidrule(lr){2-3}\cmidrule(lr){4-5}\cmidrule(lr){6-7}\cmidrule(lr){8-8}
\multicolumn{1}{c}{} &  
\multicolumn{1}{c}{N} &  
\multicolumn{1}{c}{GT} &  
\multicolumn{1}{c}{N} &  
\multicolumn{1}{c}{GT} &  
\multicolumn{1}{c}{N} &  
\multicolumn{1}{c}{GT} &  
\multicolumn{1}{c}{N/GT} \\
\midrule
misc & -0.02 & -0.21 & -0.20&-0.20 & -0.12&-0.12 & -0.17 \\
lena & 0.15&-0.25 & -0.37&-0.37 & -0.26&-0.26 & -0.30 \\
barbara & 0.05&-0.13 & -0.20&-0.20 & -0.13&-0.13 & -0.18 \\
couple & 0.12&-0.24 & -0.36&-0.36 & -0.25&-0.25 & -0.30 \\
chest & 0.25&-0.25 & -0.32&-0.07 & -0.29&-0.29 & -0.34 \\
man & 0.10&-0.22 & -0.34&-0.34 & -0.24&-0.24 & -0.27 \\
palace & 0.08&-0.20 & -0.30&-0.30 & -0.20&-0.20 & -0.24 \\
blood & 0.30&-0.27 & 0.05&0.39 & -0.28&-0.18 & -0.34 \\
f16 & 0.09&-0.20 & -0.31&-0.31 & -0.21&-0.21 & -0.26 \\
boat & 0.10&-0.22 & -0.34&-0.34 & -0.23&-0.23 & -0.27 \\
\bottomrule
\end{tabular}}
\caption{Kendall correlation values between the final PSNR scores and corresponding metrics for image denoising task. MSE metrics are calculated either with the noisy image (denoted as N) or with the ground truth image (denoted as GT).}
\label{table:dn_correlation}
\end{table}

Table \ref{table:dn_correlation} shows the correlation between the final PSNR scores and the values of the corresponding metrics for image denoising task. These correlations are extracted from the \textit{NAS Dataset for DIP}. For most of the images, the metrics calculated with ground truth (GT) correlate reasonably well with model performance. It follows that the metrics can be used to reduce the significant number of models to be trained in a NAS algorithm. However, in a practical situation, we must calculate the metrics using corrupted images since ground truth images are not accessible. In this case, the correlation of PSD DB MSE metric drops, but others still remain almost the same. The comparison between PSD DB MSE and PSD DB Strip MSE shows that focusing on specific parts of the PSDs can be used to address the correlation drop and results in higher correlation than focusing on the whole PSD. Also, images from different domains respond very differently to the metrics. For example, the correlation values for the blood and chest images are dissimilar to each other. All these observations highlight the benefits of concentrating on specific regions of PSDs rather than the whole.

\subsection{Experiments on Image Restoration Datasets}
In this section, we run experiments on commonly used datasets for image denoising, image inpainting, and single-image super-resolution tasks.  
To test our algorithm on conventional datasets, we take a subset of 5000 models from the search space and calculate the image-specific metrics for each model. For each metric, we sort the models according to their metric scores and perform selection among top-N models. Note that for the following experiments, we used the same 5000 models. In the quantitative analysis part, we describe and compare our metrics from a practical perspective. 

\noindent
\textbf{Image Restoration Set-Up.}
To compare the performance of the models selected by the proposed metrics with that of state-of-the-art work \cite{Chen2020NAS-DIP}, we evaluate on well adopted datasets. For image denoising, we use BM3D~\cite{Dabov2007VideoDB}, Set12~\cite{Zhang2017BeyondAG} and CBM3D~\cite{Dabov2007VideoDB} datasets and apply Gaussian noise with $\sigma=25$. For image inpainting, we use BM3D~\cite{Dabov2007VideoDB} and Set12~\cite{Zhang2017BeyondAG} datasets under 50\% missing pixels setting. We use Set5~\cite{Set5} and Set14~\cite{Set14} datasets for the super-resolution experiments under three upsampling scales: $2\times$, $4\times$, $8\times$. We reproduce the results of NAS-DIP~\cite{Chen2020NAS-DIP} and DIP~\cite{Ulyanov_2018_CVPR} with the code provided by the authors~\cite{Chen2020NAS-DIP, Ulyanov_2018_CVPR}, respectively. This allows a fair analysis by performing optimizations under same conditions.

\begin{table*}[t]

\centering
\begin{tabular}{lcccccc}
\toprule
\multicolumn{1}{c}{Datasets} & \multicolumn{1}{c}{DIP~\cite{Ulyanov_2018_CVPR}} & \multicolumn{1}{c}{NAS-DIP~\cite{Chen2020NAS-DIP}} &  \multicolumn{1}{c}{\begin{tabular}[c]{@{}c@{}}PSD DB \\ MSE (15\textsuperscript{n})\end{tabular}} & \multicolumn{1}{c}{\begin{tabular}[c]{@{}c@{}}PSD DB \\Strip MSE (15\textsuperscript{n})\end{tabular}} & \multicolumn{1}{c}{\begin{tabular}[c]{@{}c@{}}PSD Strip\\Hist EMD (15\textsuperscript{n})\end{tabular}} & \multicolumn{1}{c}{\begin{tabular}[c]{@{}c@{}}99\\ BW (15\textsuperscript{n})\end{tabular}} \\
 \midrule
\multicolumn{7}{c}{Denoising ($\sigma=25$)} \\ \midrule

BM3D~\cite{Dabov2007VideoDB} &\underline{27.87} & 27.44   & 24.31  & \textbf{28.39}  & 27.81  & 27.14   \\
Set12~\cite{Zhang2017BeyondAG} & \underline{27.92} & 26.88   & 23.77  & \textbf{28.06}  & 27.81 & 26.86  \\
CBM3D~\cite{Dabov2007VideoDB}& 28.93 & 29.13   & \underline{30.01}  & \textbf{30.36}  & 29.04  & 28.28 \\

\midrule
\multicolumn{7}{c}{Inpainting} \\
\midrule
BM3D~\cite{Dabov2007VideoDB} & 31.04 & 30.55 & \textbf{32.90} & 32.32 & \underline{32.75} & 30.07  \\
Set12~\cite{Zhang2017BeyondAG}& 31.00 & 30.86  & \textbf{32.22} & 31.68 & \underline{32.04} & 29.91\\
\midrule
\multicolumn{7}{c}{Super-resolution} \\
\midrule
Set5 $\times 2$~\cite{Set5}& 33.19\textsuperscript{*} & \textbf{36.16} & \underline{34.83}& 34.72 & 33.98 & 28.84  \\
Set5 $\times 4$~\cite{Set5}& 29.89\textsuperscript{*}& \textbf{30.66}  & \underline{30.22} & 30.05 & 30.03  & 25.23  \\
Set5 $\times 8$~\cite{Set5}& \underline{25.88\textsuperscript{*}} & \underline{25.88}  & 25.82 & \textbf{25.94} & 25.74 & 25.46   \\
Set14 $\times 2$~\cite{Set14}& 29.80\textsuperscript{*}  & \textbf{32.11}& 30.74 & \underline{30.87} & 29.71 & 25.63  \\
Set14 $\times 4$~\cite{Set14}& 27.00\textsuperscript{*} & \textbf{27.36}  & \underline{27.19}  & 27.03 & 27.15 & 25.03 \\
Set14 $\times 8$~\cite{Set14}& \textbf{24.15\textsuperscript{*}}& 23.96  & 23.97 & 23.90 & \underline{24.03} & 23.21  \\
\bottomrule
\end{tabular}

\caption{Final PSNR scores of DIP~\cite{Ulyanov_2018_CVPR}, NAS-DIP~\cite{Chen2020NAS-DIP}, and metrics on denoising, inpainting, and super-resolution tasks. \textsuperscript{*} denotes the PSNR scores that are evaluated at the optimal-stopping point (where we have access to ground truth). \textsuperscript{n} denotes that the \textit{full-sized averaging} selection technique is used. \textbf{Bold} denotes the highest scores, \underline{underline} denotes the 2\textsuperscript{nd} highest scores along the row. Note that super-resolution scores of DIP~\cite{Ulyanov_2018_CVPR} are directly taken from NAS-DIP paper \cite{Chen2020NAS-DIP}.}
\label{tbl:comparison}
\end{table*}

\subsubsection{Quantitative Analysis}

In Table \ref{tbl:comparison}, we report the PSNR scores of DIP~\cite{Ulyanov_2018_CVPR}, NAS-DIP~\cite{Chen2020NAS-DIP} and our metrics. We employ full-sized averaging selection among 15 models for our metrics. For denoising, PSD DB Strip MSE metric takes first place for all datasets. Also, the idea of fighting against the correlation drop that arises when using the corrupted image while calculating the metrics is consistent with the findings presented in Table \ref{tbl:comparison}. There is a large gap between PSD DB MSE and its stripped version for gray-scale datasets since eliminating the very high regions by applying a strip mask alleviates the effects of noise on the PSD, which in turn enables finding better models. Conversely, in super-resolution, they are marginally different from each other and, one possible explanation could be that our method is prone to select models with low pass characteristics. Super-resolution requires producing fine details, however our selected model produces \textit{smoother} outputs and occasionally fails to produce fine details (see Appendix), indicating the need for a different metric for super-resolution. This observation is more evident in the 99 BW metric (Table \ref{tbl:comparison}) since models with low pass characteristics are explicitly selected leading to the worst performance amongst the metrics.
For inpainting, all the image-specific metrics have outperformed both DIP~\cite{Ulyanov_2018_CVPR} and NAS-DIP~\cite{Chen2020NAS-DIP}. Overall, the metrics give promising results for denoising and inpainting tasks. Furthermore, the performance of our image independent metric (99 BW) supports the need for image-specific models since 99 BW never outperforms any one of the stripped metrics for all image restoration tasks.

\begin{table}[h]
\centering
\begin{tabular}{lcccc}
\toprule
\multicolumn{1}{c}{} & \multicolumn{2}{c}{} & \multicolumn{2}{c}{\begin{tabular}[c]{@{}c@{}}PSD DB\\ Strip MSE\end{tabular}} \\ 
\cmidrule(lr){4-5}
\multicolumn{1}{c}{Datasets}  &
\multicolumn{1}{c}{Random\textsuperscript{r}} &\multicolumn{1}{c}{NAS-DIP}  &  \multicolumn{1}{c}{15\textsuperscript{r}}& \multicolumn{1}{c}{15\textsuperscript{n}} \\
\midrule
\multicolumn{5}{c}{Denoising ($\sigma=25$)} \\ 
\midrule
BM3D~\cite{Dabov2007VideoDB}  &23.20 & 27.44 &\underline{27.66} &\textbf{28.39}\\
Set12~\cite{Zhang2017BeyondAG}  &22.66 &26.88 &\underline{27.40} &\textbf{28.06}\\
CBM3D~\cite{Dabov2007VideoDB}& 27.86&29.13 &\underline{29.28} &\textbf{30.36}\\
\midrule
\multicolumn{5}{c}{Inpainting} \\ 
\midrule
BM3D~\cite{Dabov2007VideoDB}  &32.30 &30.55 &\textbf{33.01} &\underline{32.32}\\
Set12~\cite{Zhang2017BeyondAG}  &31.64 &30.86 &\textbf{32.17} &\underline{31.68}\\
\bottomrule
\end{tabular}
\caption{2 different averaging techniques are used for PSD DB Strip MSE metric. \textsuperscript{r} denotes the resized averaging selection, \textsuperscript{n} denotes the full-sized averaging selection. ``Random" denotes the random selection over search space. \textbf{Bold} denotes the highest scores, \underline{underline} denotes the 2\textsuperscript{nd} highest scores along the row.} 
\label{table:resized}
\end{table}

\begin{figure*}[h]
\centering
\begin{subfigure}[b]{0.85\linewidth} \includegraphics[width=\linewidth]{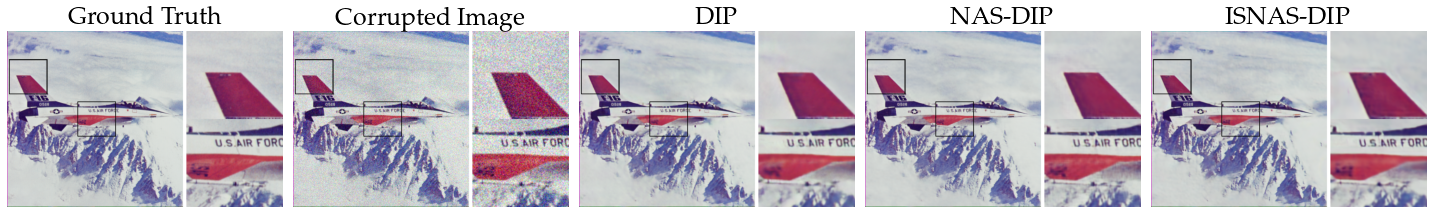}

     \end{subfigure}%
     
\centering
\begin{subfigure}[b]{0.85\linewidth} \includegraphics[width=\linewidth]{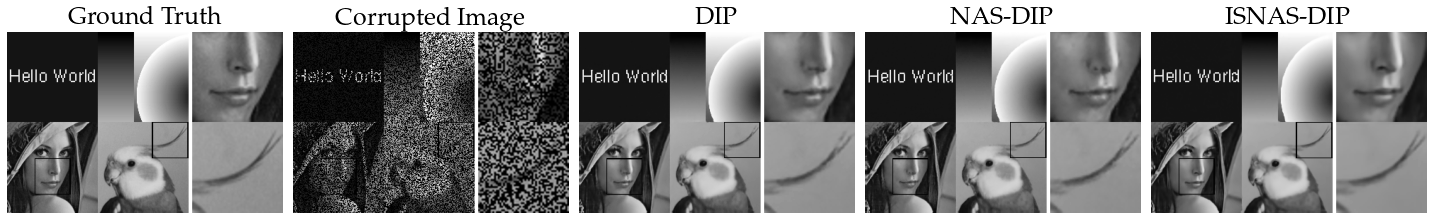}
     \end{subfigure}%

\centering
\begin{subfigure}[b]{0.85\linewidth}      \includegraphics[width=\linewidth]{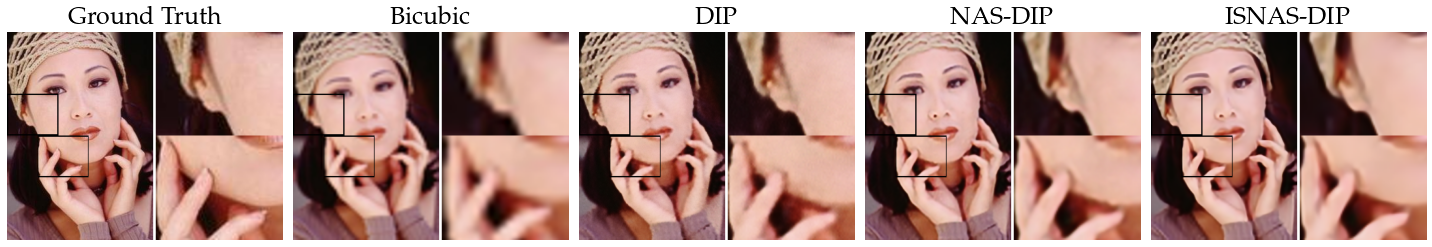}
         \label{fig:sr_qual}
     \end{subfigure}%
     \vspace*{-5mm}
    \caption{Qualitative examples of image denoising (1\textsuperscript{st} row), image inpainting (2\textsuperscript{nd} row), and single-image super-resolution (3\textsuperscript{rd} row) tasks. Each column shows the images of ground truth, corrupted image (Bicubic~\cite{Dong2016ImageSU} for super-resolution), reconstructed images of DIP~\cite{Ulyanov_2018_CVPR}, NAS-DIP~\cite{Chen2020NAS-DIP}, and ISNAS-DIP, respectively.}
    \label{fig:qual}
\end{figure*}

We further compare the performance of resized-averaging selection technique with both full-sized averaging selection and a random baseline. As the random baseline, 15 models are selected randomly from the search space, then resized-averaging is applied to perform the selection. Table~\ref{table:resized} shows the final PSNR scores of the random baseline, NAS-DIP~\cite{Chen2020NAS-DIP}, and PSD DB Strip MSE metric evaluated under two selection procedures. It demonstrates that PSD DB Strip MSE metric performs significantly better than random selection. Additionally, owing to the image-specific model need, it performs better for all datasets.

There is a tradeoff between the speed and the performance of averaging techniques, especially in image denoising. For a $256\times256$ grayscale image, the optimization speed is approximately 5$\sim$8 it/s whereas it rises to $\sim$20 it/s when the image is resized to $64\times64$. For a selection among 15 models for denoising, it requires approximately 15 minutes for resized-averaging and 37.5$\sim$60 minutes for full-sized averaging technique. These cases are much faster than NAS-DIP~\cite{Chen2020NAS-DIP} selection since the authors claim that the search lasted 3 days for denoising in total.

We note that the reason we compared with NAS-DIP in our experiments is that they have outperformed previous learning-free DIP approaches. 

\subsubsection{Qualitative Analysis}

Figure \ref{fig:qual} shows qualitative samples of several images for all tasks. Generally, we observe apparent visual improvements in the quality of the restored image. For example, in denoising, fine details such as the notch on top of the fin of F16 are retrieved better with ISNAS-DIP. However, in super-resolution, since NAS-DIP~\cite{Chen2020NAS-DIP} performs better, we do not observe much improvement. The improvements in denoising and inpainting emphasize the importance of taking into consideration the content of an input image while designing network architectures.



\section{Conclusion, Limitations and Future Work}

In this work we show that the optimal architecture to exploit the deep image prior is image dependent. Based on this insight, we propose several metrics that allow for a fast image-specific neural network architecture search and show that the found models outperform benchmark architectures for image restoration tasks using DIP. Furthermore, we establish a \textit{NAS Dataset for DIP} that can be used for future research investigating image-specific NAS.


There are still some limitations that future work entails. We show that for N=15, our metrics are able to select a subset that contains optimal image-specific models. That is, we narrow down the whole search space to just 15 models. However, we are still limited in selecting the top-1 model. The current averaging technique needs N models to be optimized. We attempt to boost the optimization process by using the resized image while optimizing 15 models, which is shown to expedite the process. Yet, it does not guarantee that the selected one is the best among this subset, but is the one that is close to the average. Hence, selecting the top-1 model among a subset of models is still an open question.

One other issue is the early stopping problem of DIP. In our experiments, we use a fixed number of iterations for each restoration task. We observe a significant correlation between the proposed metrics and the optimal number of iterations for each image (see Appendix). This means that the metrics might also be useful to determine the early stopping point for the training of selected models.

Moreover, the \textit{NAS Dataset for DIP} experiments reveal the need for better-suited metrics for different types of images since the domains of images influence the correlations. Future work entails implementing \textit{learned} metrics rather than hand-designed definitions as in our case. A deep learning model can be utilized in order to do so. Using a search space, a metric can be described by a \textit{neural network}, which can be trained with the CNNs' outputs and corrupted images to obtain a better formulation of a metric.


{\small
\bibliographystyle{unsrt}
\bibliography{main}
}
\onecolumn
\begin{appendices}

\section{Overview}
This appendix provides details and results in addition to the main manuscript. Furthermore, codes and dataset are available at \href{https://github.com/ozgurkara99/ISNAS-DIP}{https://github.com/ozgurkara99/ISNAS-DIP}.

\section{Deep Image Prior}
In deep image prior, the network $f_{\theta}$ is enforced to map the noise $z$ to the uncorrupted version of the corrupted image $x_0$ by minimizing the following objective functions with respect to $\theta$ depending on the task of interest, and stopping the optimization at a pre-determined iteration point to prevent overfitting. We follow DIP~\cite{Ulyanov_2018_CVPR} and optimize the following objectives:
\begin{align}
    \mathcal{L}_{denoising}(\theta) = ||f_{\theta}(z) - x_0||^2
\end{align}
\begin{align}
    \mathcal{L}_{inpainting}(\theta) = ||(f_{\theta}(z) - x_0) \otimes M||^2
\end{align}
\begin{align}
    \mathcal{L}_{superresolution}(\theta) = ||D(f_{\theta}(z)) - x_0||^2
\end{align}
where $D(\cdot)$ denotes the downsampling operator, $\otimes$ denotes pixel-wise multiplication, and $M$ denotes the mask for inpainting.

\section{Exponential Averaging}

Inspired by DIP \cite{Ulyanov_2018_CVPR}, we take the average of the reconstructed predictions $x_t$ as our final image. Unlike DIP~\cite{Ulyanov_2018_CVPR} and NAS-DIP~\cite{Chen2020NAS-DIP}, we did the averaging not only for image denoising but also for other tasks as well since we saw that it substantially improves the quality in all tasks. Note that we used exponential averaging for all approaches, i.e. DIP~\cite{Ulyanov_2018_CVPR}, NAS-DIP~\cite{Chen2020NAS-DIP} and ISNAS-DIP. It is formulated as follows:
\begin{align}
    x^* = \gamma^{(T-1)} \cdot x_1 + \sum_{t=2}^{T} x_t \cdot \gamma^{T-t} \cdot (1-\gamma)
\end{align}
where $x^*$ denotes the final result, T denotes the total iteration number, $x_t$ denotes the restored output at $t^{th}$ iteration, and $\gamma$ is selected to be 0.99.

\section{Architecture Selection}
\begin{figure}[H]
\centering
\includegraphics[width=0.75\linewidth]{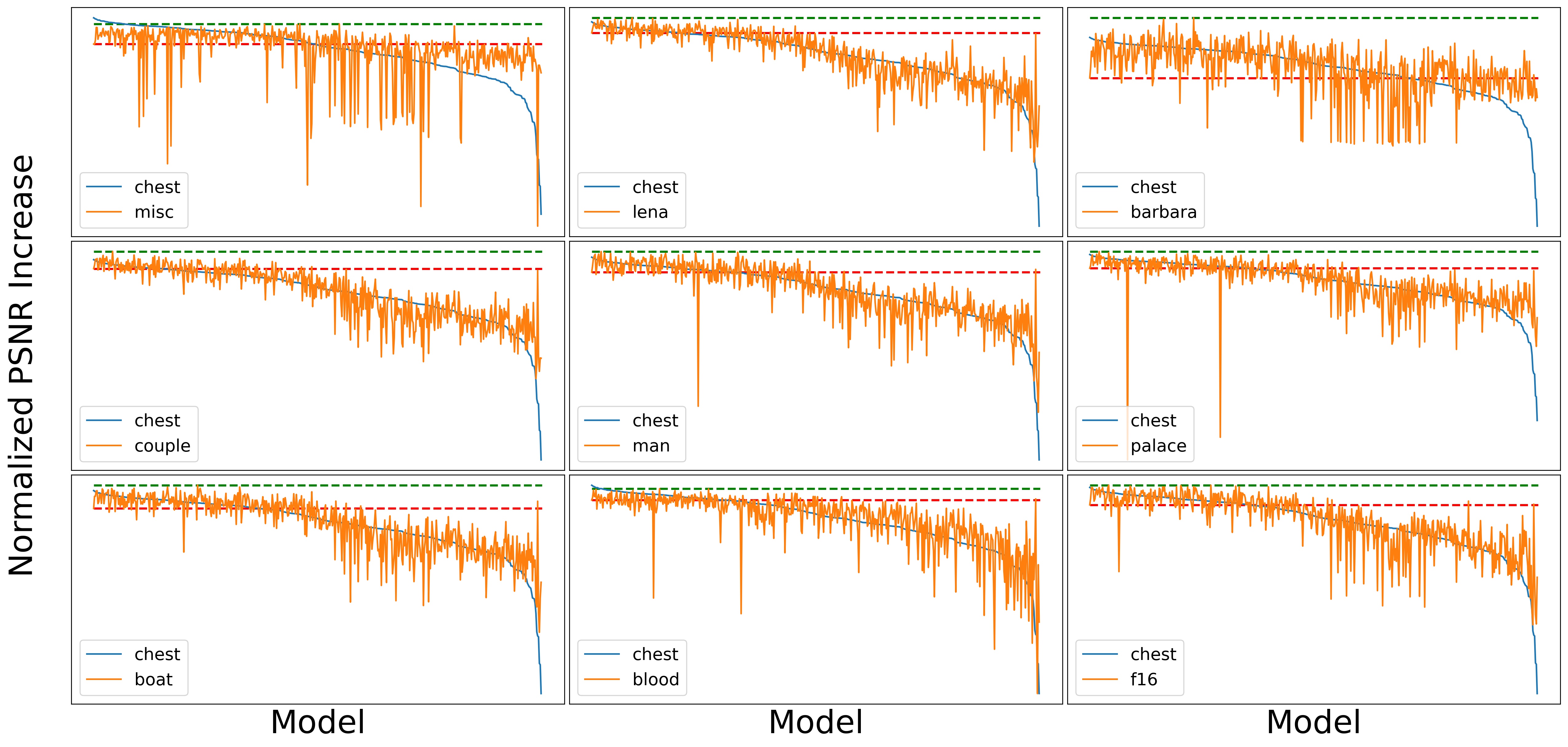}
\caption{Normalized PSNR increase scores for each model in \textit{NAS Dataset}. Each subplot corresponds to the scores of models that are sorted in descending order according to their scores on the ``reference image" (\textcolor{blue}{blue} line) and evaluated on the ``target image" (\textcolor{orange}{orange} line). \textcolor{red}{Red} dashed line depicts the normalized PSNR increase score of the best model of reference image evaluated on target image. \textcolor{green}{Green} dashed line shows the normalized PSNR increase score of the best model of the target image}
\label{fig:mot}
\end{figure}

We plot the normalized\footnote{We normalized the data by subtracting mean and dividing the deviation as follows: $x_{normalized} = \frac{x - mean}{std}$} PSNR increase scores for each model in \textit{NAS Dataset for DIP} as shown in Fig. \ref{fig:mot}. Each subplot represents normalized PSNR increase scores for different image pairs. Models are sorted according to their normalized PSNR increase scores measured for the reference image (depicted with \textcolor{blue}{blue} line) and evaluated on the target image (depicted with \textcolor{orange}{orange} line).
We select ``chest" as our reference image since it belongs to a different domain as opposed to other images. 

As can be seen in Fig.~\ref{fig:mot}, although the trend of the orange line is similar to that of reference image apparently; same models tend to perform differently. Furthermore, the worst models for the reference image may perform better on the target image. 

The gap between green and red dashed line show how much the performance differs between best model for reference image and best model of the target image both evaluated on target image. For instance, we observe a large gap between the dashed lines for chest and barbara images due to the difference in domains that each belongs to.

\section{Correlation between optimal stopping point}
We observed that different models within the same search space have different optimal stopping points as shown in Fig.~\ref{fig:stop}. We questioned if the optimal stopping point is an architectural property or not. In light of these questions, we report the correlations between our metrics and optimal stopping points as represented in Table~\ref{table:iter_corr}. Surprisingly, there is a strong correlation between our metrics and the optimal stopping point. Our work will be extended by further investigation of the detection of optimal stopping point utilizing the proven strong correlation.

  \begin{minipage}{\textwidth}
  \begin{minipage}[b]{0.49\textwidth}

\includegraphics[width=\linewidth]{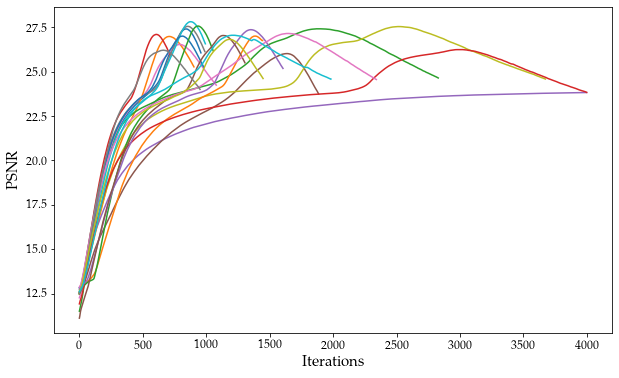}
\captionof{figure}{PSNR curve of different models optimized for the ``barbara" image}
\label{fig:stop}

  \end{minipage}
  \hfill
  \begin{minipage}[b]{0.49\textwidth}

\centering
\resizebox{\linewidth}{!}{
\begin{tabular}{lccccccc}
\toprule
\multicolumn{1}{c}{Images} &  \multicolumn{2}{c}{\begin{tabular}[c]{@{}c@{}}PSD DB \\ MSE\end{tabular}} & \multicolumn{2}{c}{\begin{tabular}[c]{@{}c@{}}PSD DB \\ STRIP MSE\end{tabular}} & \multicolumn{2}{c}{\begin{tabular}[c]{@{}c@{}}PSD STRIP\\ HIST EMD\end{tabular}} & \multicolumn{1}{c}{\begin{tabular}[c]{@{}c@{}}PSD 99 PER\\ BW\end{tabular}} \\ 
\cmidrule(lr){2-3}\cmidrule(lr){4-5}\cmidrule(lr){6-7}\cmidrule(lr){8-8}
\multicolumn{1}{c}{} &  
\multicolumn{1}{c}{N} &  
\multicolumn{1}{c}{GT} &  
\multicolumn{1}{c}{N} &  
\multicolumn{1}{c}{GT} &  
\multicolumn{1}{c}{N} &  
\multicolumn{1}{c}{GT} &  
\multicolumn{1}{c}{N/GT} \\
\midrule
misc & 0.24 & -0.06 & -0.45 & -0.45 & -0.44 & -0.44 & -0.39 \\
lena & 0.30 & -0.40 & -0.41 & -0.41 & -0.43 & -0.43 & -0.38 \\
barbara & 0.33 & -0.43 & -0.46 & -0.46 & -0.47 & -0.47 & -0.41 \\
couple & 0.28 & -0.39 & -0.40 & -0.40 & -0.43 & -0.44 & -0.36 \\
chest & 0.32 & -0.36 & -0.19 & 0.15 & -0.40 & -0.40 & -0.34 \\
man & 0.28 & -0.40 & -0.41 & -0.42 & -0.44 & -0.44  & -0.37 \\
palace & 0.27 & -0.35 & -0.36 & -0.37 & -0.40 & -0.40 & -0.34 \\
blood & 0.20 & -0.22 & 0.20 & 0.17 & -0.06 & 0.04 & -0.18 \\
f16 & 0.28 & -0.41 & -0.42 & -0.42 & -0.43  & -0.43 & -0.38 \\
boat & 0.30 & -0.41 & -0.40 & -0.40 & -0.43 & -0.43 & -0.37 \\
\bottomrule
\end{tabular}}
\captionof{table}{Kendall correlation values between the optimal stopping points and corresponding metrics for image denoising task. MSE metrics are calculated either with the noisy image (denoted as N) or with the ground truth image (denoted as GT).}
\label{table:iter_corr}
    \end{minipage}
  \end{minipage}

\section{Histograms}
Fig. \ref{fig:histogram} shows the distribution of the final PSNR scores of the 522 models for both image denoising (a) and image inpainting (b). Note that the best-performing model found with PSD DB Strip MSE metric is shown with \textcolor{red}{red} for N=15 and black for N=5 case.

\begin{figure}[ht]
\centering
\begin{subfigure}[b]{\linewidth}
     \includegraphics[width=\textwidth]{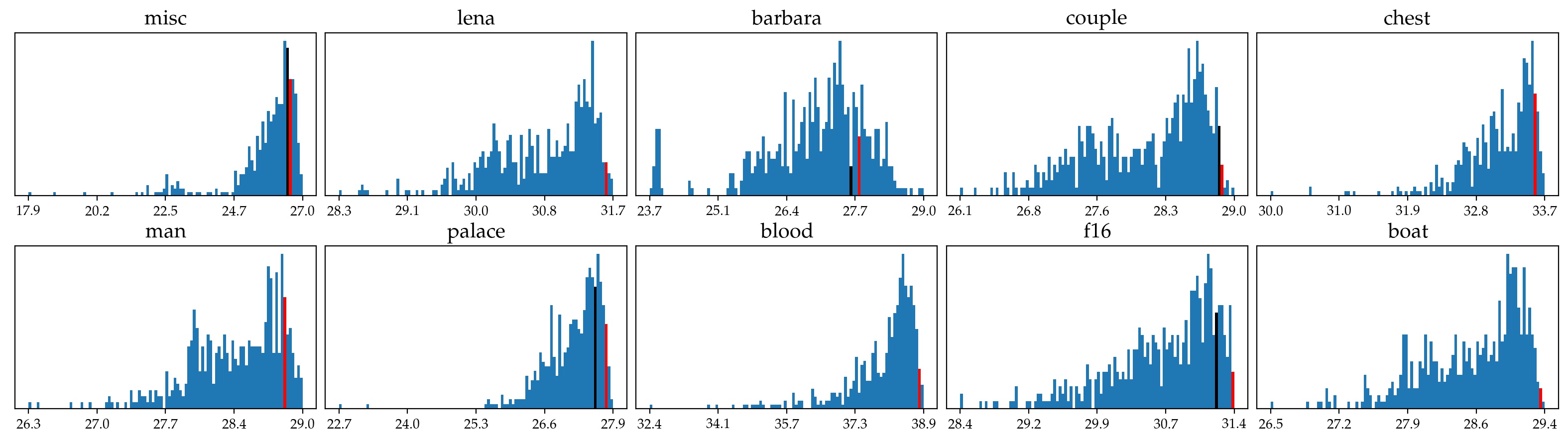}
     \caption{Denoising}
     \label{fig:psd_db_mse}
    \end{subfigure}%
    
\begin{subfigure}[b]{\linewidth}
     \includegraphics[width=\textwidth]{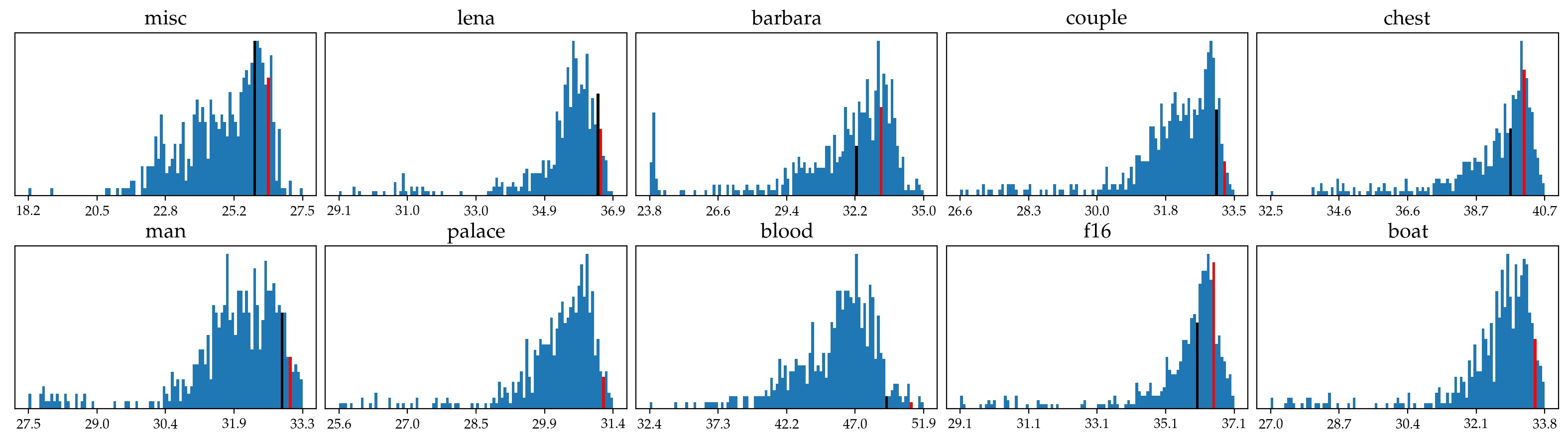}
     \caption{Inpainting}
     \label{fig:psd_strip_emd}
\end{subfigure}%
\caption{Histograms of the final PSNR scores of models for all images in \textit{the NAS Dataset for DIP} for image denoising and image inpainting . Non-blue colored bars represent the final PSNR scores of the best models included in top-N models found by PSD DB Strip MSE metric. The bin depicted with \textcolor{red}{red} denotes the best model among N=15 models, the black bin denotes the best one among N=5 models. }
\label{fig:histogram}
\end{figure}

\clearpage
\section{Qualitative Analysis}

\begin{figure*}[h]
\begin{subfigure}[b]{\linewidth} \includegraphics[width=\linewidth]{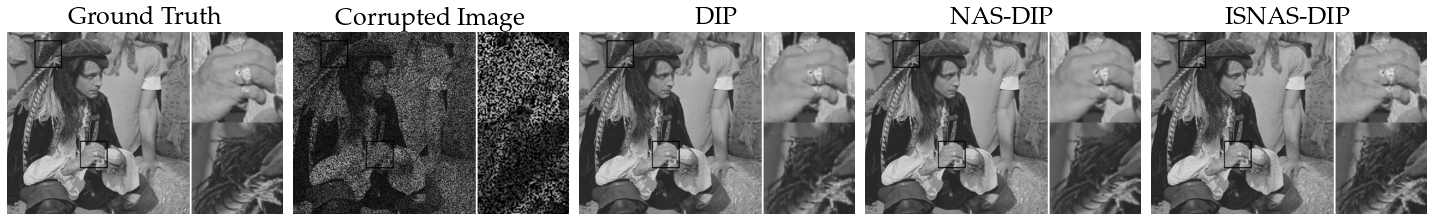}
     \end{subfigure}%
     
    \begin{subfigure}[b]{\linewidth}  \includegraphics[width=\linewidth]{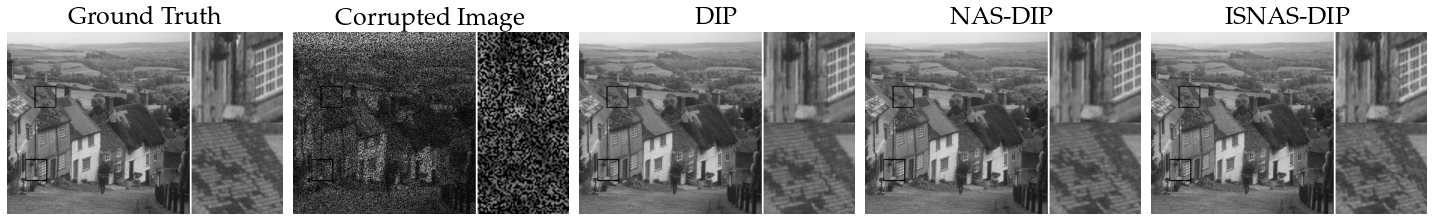}
         \caption{Inpainting}
     \end{subfigure}%
     
      \centering
    \begin{subfigure}[b]{\linewidth}         \includegraphics[width=\linewidth]{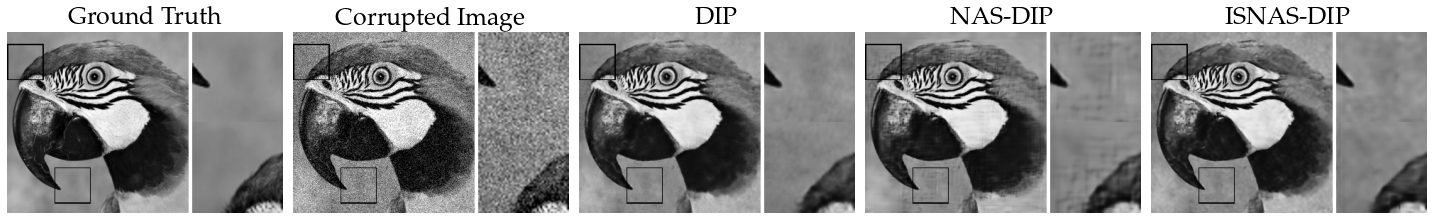}
    \end{subfigure}
     \centering
    \begin{subfigure}[b]{\linewidth}         \includegraphics[width=\linewidth]{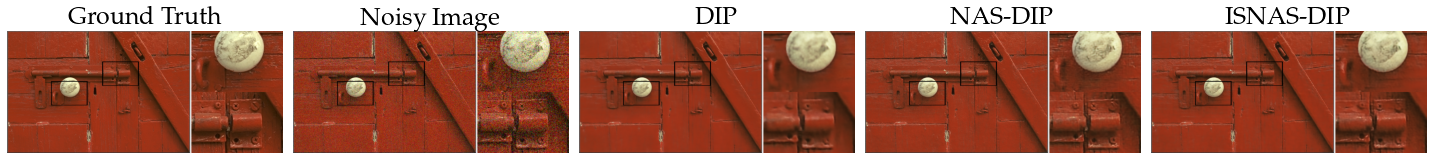}
         \caption{Denoising}
         \label{fig:sr_qual}
     \end{subfigure}%
    \caption{Qualitative examples of image inpainting (a), and image denoising (b) tasks. Each column shows the images of ground truth, corrupted image, reconstructed images of DIP~\cite{Ulyanov_2018_CVPR}, NAS-DIP~\cite{Chen2020NAS-DIP}, and ISNAS-DIP, respectively.}
    \label{fig:qual2}
\end{figure*}

We show 2 more qualitative examples for denoising and inpainting tasks in Fig.~\ref{fig:qual2}.

\begin{figure}[ht]
\centering
\begin{subfigure}[b]{0.3\linewidth}
     \includegraphics[width=\textwidth]{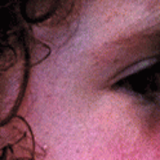}
     \caption{Ground Truth}
    \end{subfigure}%
\hspace{20pt}
\centering
\begin{subfigure}[b]{0.3\linewidth}
     \includegraphics[width=\textwidth]{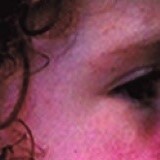}
     \caption{NAS-DIP~\cite{Chen2020NAS-DIP}}
    \end{subfigure}%
\hspace{20pt}
\centering
\begin{subfigure}[b]{0.3\linewidth}
     \includegraphics[width=\textwidth]{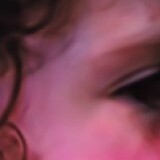}
     \caption{ISNAS-DIP}
\end{subfigure}%
\caption{A qualitative example for super-resolution $\times 2$}
\label{fig:superres_qual}
\end{figure}

\section{Super-resolution Results}
Fig.~\ref{fig:superres_qual} shows the performance of NAS-DIP~\cite{Chen2020NAS-DIP} and our method on a super-resolution image with $\times2$ scale. It is evident in the example that our method is not good at producing finer details since it tends to select models having more low-pass characteristics. Hence, it is not able to outperform previous approaches, indicating the need for a different metric for super-resolution.

\section{Histogram Analysis with Random Selection}

\begin{figure*}[h]

\centering
\includegraphics[width=1\textwidth]{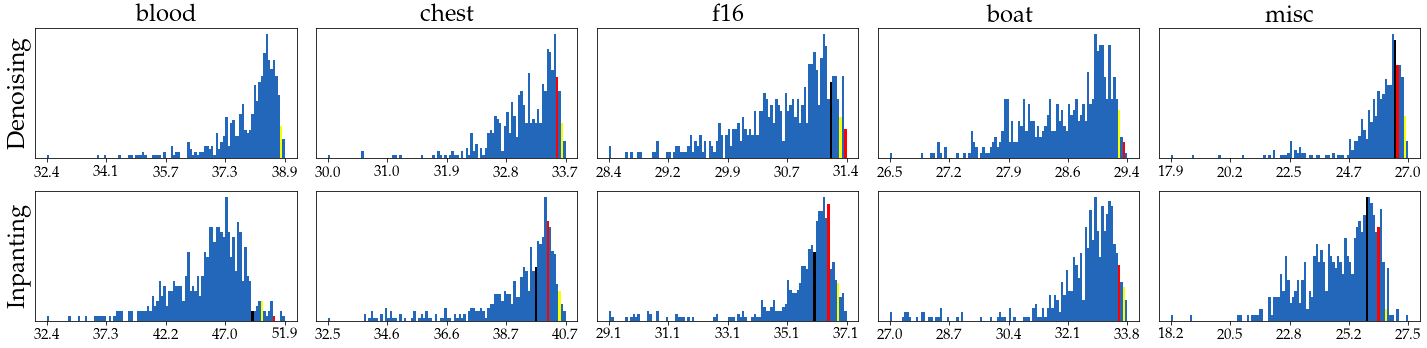}

\caption{Histograms of the final PSNR scores of models for all images in \textit{the NAS Dataset for DIP} for image denoising and image inpainting . \textcolor{red}{Red} and black colored bars represent the final PSNR scores of the best models included in top-N models found by PSD DB Strip MSE metric. The bin depicted with \textcolor{red}{red} denotes the best model among N=15 models, the black bin denotes the best one among N=5 models. The bin depicted with \textcolor{yellow}{yellow} denotes the average PSNR scores (which was repeated 10 times) of the models each of which was the best performing model among randomly sampled 15 models.}
\label{fig:hist_rand}
\end{figure*}

In Fig.~\ref{fig:hist_rand}, we additionally provide the resulting PSNR of the best performing models among a random selected 15 models 10 times, which is shown with \textcolor{yellow}{yellow} bin. In other words, at each time, we select 15 different models randomly, and record the best performing model's PSNR among these, and repeat this procedure 10 times to take the average. Whereas this is similar to the PSNR of the best model selected by the PSD DB Strip MSE metric it should be noted that it is not a deterministic selection process in contrast to our proposed method and can thus fail unexpectedly. Thus, in order to make a fair comparison, the variances should be noted.

\end{appendices}

\end{document}